\documentclass[letterpaper, 10 pt, conference]{conf/ieeeconf}  

\IEEEoverridecommandlockouts                              

\usepackage{cite}
\usepackage{amsmath,amssymb,amsfonts}
\usepackage{algorithmic}
\usepackage{graphicx}
\usepackage{textcomp}
\usepackage{xcolor}
\usepackage{bm}
\usepackage{gensymb}
\usepackage{subfigure}
\usepackage{balance}

\DeclareMathOperator{\atan}{atan2}

\usepackage{url}
\usepackage[linkcolor=black,citecolor=black,urlcolor=black,colorlinks=true]{hyperref}

\title{\LARGE \bf
Air Bumper: A Collision Detection and Reaction Framework for Autonomous MAV Navigation
}

\author{Ruoyu Wang, Zixuan Guo, Yizhou Chen, Xinyi Wang, Ben M. Chen
\thanks{The work was supported in part by the Research Grants Council of Hong Kong SAR under Grants 14209020 and 14206821, and in part by the InnoHK of the Government of Hong Kong via the Hong Kong Centre for Logistics Robotics. Authors are with the Chinese University of Hong Kong, Shatin, N.T., Hong Kong 999077. (Email: \{rywang, zxguo, josephchen, xywangmae\}@link.cuhk.edu.hk, bmchen@cuhk.edu.hk).}
}

\begin{document}

\maketitle
\thispagestyle{empty}
\pagestyle{empty}

\begin{abstract}

Autonomous navigation in unknown environments with obstacles remains challenging for micro aerial vehicles (MAVs) due to their limited onboard computing and sensing resources. Although various collision avoidance methods have been developed, it is still possible for drones to collide with unobserved obstacles due to unpredictable disturbances, sensor limitations, and control uncertainty.
Instead of completely avoiding collisions, this article proposes Air Bumper, a collision detection and reaction framework, for fully autonomous flight in 3D environments to improve the safety of drones. Our framework only utilizes the onboard inertial measurement unit (IMU) to detect and estimate collisions. We further design a collision recovery control for rapid recovery and collision-aware mapping to integrate collision information into general LiDAR-based sensing and planning frameworks. Our simulation and experimental results show that the quadrotor can rapidly detect, estimate, and recover from collisions with obstacles in 3D space and continue the flight smoothly with the help of the collision-aware map. Our Air Bumper will be released as open-source software on GitHub\footnote[1]{\label{ftnt:airbumper}\url{https://github.com/ryrobotics/air_bumper}}.

\end{abstract}

\section{Introduction}
MAVs have gained increasing popularity for their ability to access and operate in environments that are difficult or impossible for humans to reach, making them valuable tools in various fields like infrastructure inspection \cite{rakha2018review, chan2015towards, montero2015past}, subterranean exploration \cite{tranzatto2022cerberus, agha2021nebula, hudson2021heterogeneous}, and search and rescue \cite{bi2019lightweight, horyna2023decentralized}, etc. However, safety becomes a critical concern for MAVs when operating in such complex and cluttered environments. These scenarios present a significant challenge for MAVs to conduct safe and collision-free flights. To address this challenge, much research has focused on utilizing onboard sensors such as LiDAR \cite{xu2022fast}, stereo cameras, and RGB-D cameras \cite{campos2021orb} for Simultaneous Localization and Mapping (SLAM); motion planning algorithms \cite{zhou2021raptor, lai2019model} have been developed to generate collision-free paths. Despite these efforts, MAVs are still susceptible to colliding with obstacles due to unpredictable disturbances, sensor limitations, and control uncertainty.

\begin{figure}[!htb]
    \vspace{6pt}
    \centering
    \includegraphics[width=1.0 \linewidth]{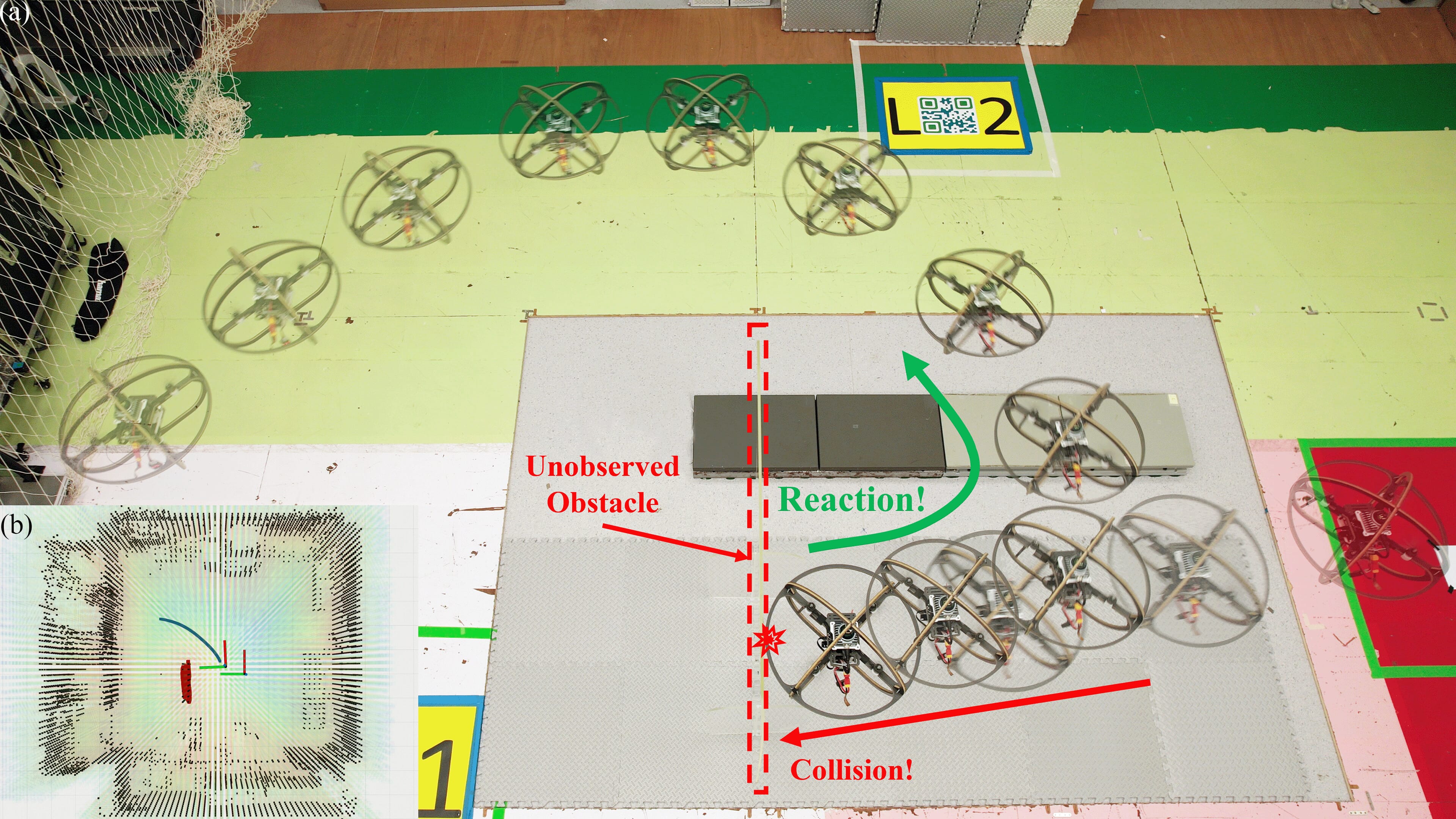}
    \caption{A collision detection and reaction experiment with an unobserved obstacle. (a) Composite images of the experiment. (b) Collision-aware volumetric map with collision point cloud.}
    \label{fig:cage_uav}
    \vspace{-12pt}
\end{figure}
Instead of dealing with MAV collision by completely avoiding it, increasing attention has been shifted to collision detection and reaction. In this paper, we introduce a unified IMU-based collision detection and reaction framework (Air Bumper) that estimates collisions and integrates the collision information into a general autonomous MAV navigation framework. To handle collisions effectively, a collision-aware volumetric mapping algorithm is developed, which collaborates with general motion planning algorithms to enable the MAVs to reach their original targets without getting stuck by obstacles. Notably, the collision detection and estimation only rely on IMU data from the flight controller without requiring any external sensors. Moreover, a fully autonomous collision-resilient MAV with a 3D cage is designed, crafted, and evaluated. This MAV itself is effectively tolerant of collisions, and its collision resilience and autonomy can be further enhanced by incorporating the proposed framework, along with general autopilot, SLAM, and motion planning algorithms. The framework enables the drone to detect and react to unobserved collisions, as well as update a collision-aware map for autonomous navigation after collisions (Fig. \ref{fig:cage_uav}). The experiments conducted in simulated and real unknown environments demonstrate that our proposed framework effectively facilitates MAV recovery from collisions with transparent and unpredictable obstacles in 3D spaces, allowing them to continue their assigned flight tasks.



\begin{figure}[t]
    \vspace{6pt}
    \centering
    \includegraphics[width=1.0 \linewidth]{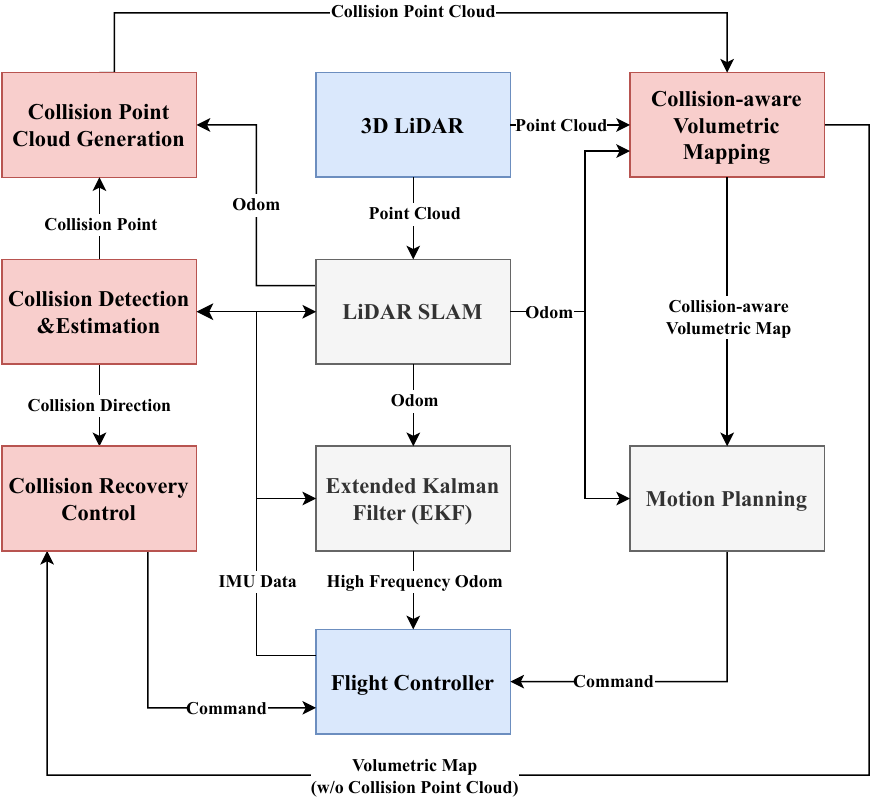}
    \caption{Overview of the collision detection and reaction framework.}
    \vspace{-6pt}
    \label{fig: workflow}
    \vspace{-12pt}
\end{figure}

\section{Related Works}
\label{Related Works}
In the face of possible collisions in flight, many researchers choose not to generate a collision-free path to avoid the collision but to design collision-resilient MAVs to deal with it. At the hardware level, there are many kinds of designs and structures to enhance collision resilience. As a high-speed rotating part, the propeller is the most vulnerable to damage in a collision. Therefore, propeller guards \cite{salaan2019development, liu2021toward, wang2020fly} are commonly used to protect it. At the same time, many cage-like structures are designed to provide more protection for the whole drone. Rigid cage structure \cite{de2022rmf, gao2023uav} can use its strength to protect inside fragile parts, like sensors, flight controllers, and onboard computers.

In addition to minimizing the impact of collisions through the hardware design discussed above, some researchers are also extracting environmental information from collisions in order to integrate it into the MAV perception system. Lew et al. in \cite{lew2019contact} proposed a contact-based inertial odometry (CIO), which can provide a usable but inaccurate velocity estimation for a hybrid ground and aerial vehicle performing autonomous navigation. In the flight, several not destructive collisions happen, and the controller can get updated information from collisions. The work in \cite{mulgaonkar2020tiercel} analyzes the impact of collisions on visual-inertial odometry (VIO) and uses collision information to build a map with a downward camera for localization. In their experiment, two glass walls are included to present that the transparent objects may cause LiDAR to get an inaccurate distance. Still, collision mapping can help MAVs detect these transparent walls. Authors in \cite{liu2021toward, lu2022online} introduce hall sensors to detect collisions and estimate the intensity and location of the collision to realize reaction control.

However, these works tend to navigate using only IMU or directly use collision data to perform reaction control, which makes the collision information hard to be recorded and reused. Although the method proposed in \cite{wang2020fly} successfully achieves collision recording for further flight in a laboratory environment using motion capture systems, the lack of integration with online sensing and planning modules limits its applicability in real-world settings. Additionally, most of these works \cite{dicker2017quadrotor, dicker2018recovery} focus on collision detection and characterization in a 2D environment. However, the obstacles in cluttered environments are often not on the same level as MAVs, which means that collisions can occur from any direction. In this work, we combine the Air Bumper framework with LiDAR-based sensing on a caged, collision-resilient MAV. This allows for collision detection and estimation in 3D space and the generation of smooth reaction trajectories with the help of collision-aware mapping.


\section{System Overview}
\label{System Overview}

\subsection{Overview of Air Bumper Framework}
The structure of our proposed collision detection and reaction framework, Air Bumper, is shown in Fig. \ref{fig: workflow}. When an MAV is flying in unknown environments, it may collide with obstacles due to the onboard sensors' limitations. In this condition, the collision detection and estimation module of our framework will use inertial data from the flight controller to estimate the collision points and feed the collision information into collision reaction modules. In collision reaction parts, the collision recovery control algorithm will utilize the direction of the collision point and the known obstacle information to command the MAV away from obstacles. Meanwhile, it will also generate a collision point cloud to the collision-aware mapping module so that the position of unobserved obstacles can be stored for further navigation. Using the updated collision-aware map, a general motion planning system can easily get the ability to deal with unobserved obstacles in 3D environments.

\subsection{Design of Collision-Resilient MAV}
\label{Design of Collision-Resilient MAV}

\begin{figure}[h]
    \centering
    \includegraphics[width=1.0 \linewidth]{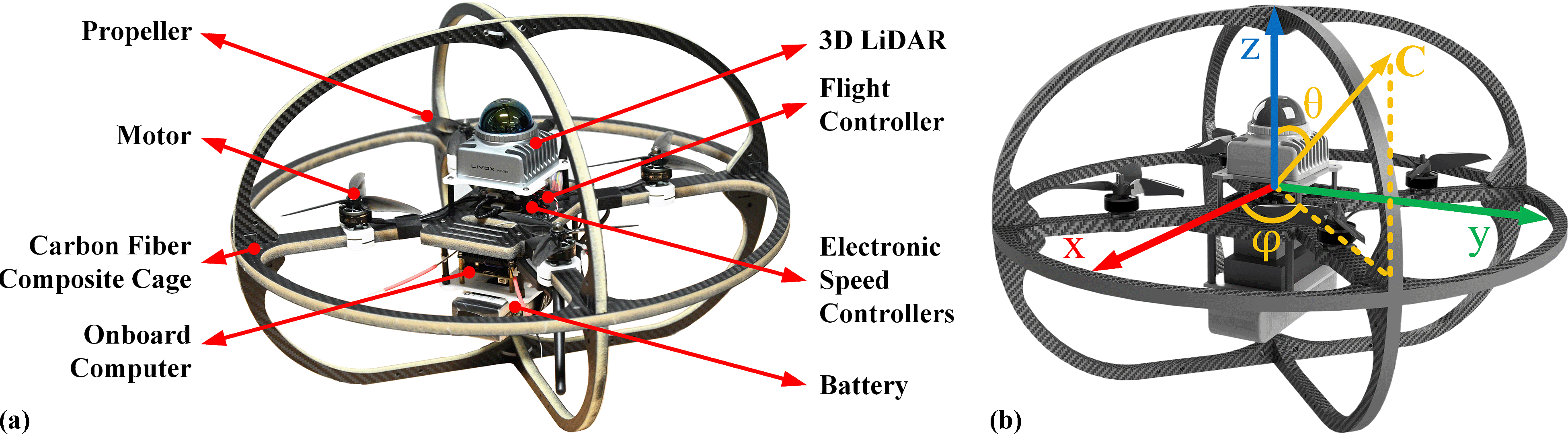}
    \caption{(a) Overview of the collision-resilient MAV design. (b) Demonstration of the intensity and direction of a collision $\bm{C}$.}
    \label{fig: hardware}
    \vspace{-6pt}
\end{figure}

The collision resilient MAV (Fig. \ref{fig: hardware}(a)) is constructed from a composite material of carbon fiber and PVC foam \cite{de2022rmf, gao2023uav}, 3D printed parts, and commercial electrical component, weighing 1.45 kg with a battery and 3D LiDAR. The dimensions of the MAV are $46 \times 46 \times 30$ cm ($\text{L} \times \text{W} \times \text{H}$). The lightweight frame and cage, crafted from composite materials, offer comprehensive protection for onboard components, and the circular design enhances the efficiency and accuracy of collision estimation. Livox Mid-360 LiDAR sensor has been selected to enable 360\degree \ of the horizontal field of view (FOV) and 59\degree \ of vertical FOV for autonomous navigation. A flight controller, Kakute H7, with PX4 autopilot, is utilized for low-level control. NVIDIA Xavier NX module with ROS framework is chosen as the onboard computer, which provides computing capabilities for Air Bumper, LiDAR SLAM \cite{xu2022fast}, GPU-accelerated volumetric mapping \cite{chen2022gpu}, and motion planning \cite{xi2021gto} algorithms.




\section{IMU-based Collision Detection and Estimation in 3D Space}
\label{Collision Detection and Estimation}
\subsection{Collision Detection}
To make Air Bumper easier to implement on any platform, IMU, the most common drone sensor, is used to collect linear acceleration data on $x$, $y$, and $z$ axes to detect the collisions rapidly. When the collision happens, the contact force will cause an additional acceleration on the MAV, and the measured value on the corresponding axes will significantly differ from the normal state. Different from the previous work \cite{mulgaonkar2020tiercel, wang2020fly, liu2021toward}, they only consider detecting the collision on a horizontal plane or using acceleration data on the $z$-axis to assist the horizontal detection. Our method also takes into account collisions other than those from horizontal planes. This feature can assist popular caged MAVs in detecting collisions from any angle. Let $\bm{a}$ as the acceleration vector of the MAV in the body frame.


Based on the analysis of acceleration data, we found that an acceleration sample can be identified as a potential collision signal if the magnitude of its component on either the $x$ or $y$ axis, represented as $|{a}_x|$ or $|{a}_y|$, exceeds an experimentally determined threshold, denoted as $a^*$. The relative threshold for collision detection on the $z$-axis is also $a^*$, but the gravity constant $g=9.81\ m/s^2$ must be considered. Meanwhile, the impact of a collision on MAV may cause several related abnormal acceleration data samples. To realize robust collision detection and estimation, a sliding-window method is used to select the maximum value from $N$ samples following the first acceleration data that exceeds the threshold. After the selection, the most represented data sample for one collision will be recorded. During the collision detection stage, the threshold ($a^*$) and sliding window size ($N$) are the only parameters that need to be tuned to filter out sensor noise and post-impact of the collision, which can be easily adjusted according to specific hardware.

\subsection{Collision Estimation}

The collision estimation module estimates the intensity and direction of collision $\bm{C}$ in the body frame (see Fig. \ref{fig: hardware}(b)) for collision recovery control. It also outputs a collision point $\bm{p}_{c}$ in the body frame for generating corresponding collision point clouds. Firstly, we need to compute a collision acceleration vector $\bm a_c$ in the body frame. This vector is directed opposite to the measured MAV acceleration vector $\bm{a}$. Therefore, we can establish the following relationships: $a_{c,x} = -a_{x},\ a_{c,y} = -a_{y},\ a_{c,z} = -(a_{z}-\bm{R}g\bm{e_z})$, where $\bm{R}$ is the rotation matrix from the world frame to body frame, $\bm{e_z}$ is the unit vector along $z$-axis. Then, the collision intensity $C$ and collision direction $\hat{\bm{C}}$ can be calculated as follows:
\begin{gather}
    C = \sqrt{{a_{c,x}}^2 + {a_{c,y}}^2 + {a_{c,z}}^2} \\
    \hat{\bm{C}} = [\sin\theta\cos\varphi, \sin\theta\sin\varphi, \cos\theta]^T
\end{gather}
Here the angles $\varphi$ and $\theta$ are computed as:
\begin{gather}
    \theta = \atan (\sqrt{{a_{c,x}}^2 + {a_{c,y}}^2},\ a_{c,z}) \\
    \varphi = \atan (a_{c,y},\ a_{c,x})
\end{gather}
where $\varphi \in (-\pi, \pi]$ is the azimuth angle and $\theta \in [0,\pi]$ is the polar angle. 

With the collision intensity and direction, we can now estimate the collision point $\bm{c}_{c}$, which is typically located on the edge of the protective cage where the MAV and an obstacle are most likely to collide. Estimating this collision point is key for updating a collision-aware map, consequently facilitating autonomous navigation. For computational simplicity, we assume that the cage of our drone is a sphere with a radius $l$. Then, the collision point can be estimated by:
\begin{gather}
    \bm{c}_{c} = \hat{\bm{C}} \cdot l
\end{gather}
%


\section{Collision Reaction in 3D Space}
\label{Collision Reaction}
Our collision reaction method aims first to utilize a straightforward but rapid recovery control strategy to quickly guide the MAV away from obstacles and restore its stability. Then, the mapping-related modules transfer the collision point into a corresponding collision point cloud and integrate it with a volumetric mapping algorithm. This enables the robot to record the estimated positions of obstacles in the world frame and navigate to the pre-collision goal using general motion planning algorithms.

\subsection{Collision Recovery Control Strategy}

When a collision occurs, an MAV without our framework will attempt to maintain its target velocity but will fail to achieve it. The motors will persist in trying to accelerate, causing the MAV to continuously collide with the obstacle and ultimately crash. To tackle this issue, we introduce a collision recovery control strategy. In addition to the basic equilibrium bounce reaction method \cite{tomic2017external, bui2022tombo}, which generates a reaction position opposite to the collision direction, our strategy aims to concurrently consider environmental information. By incorporating environmental information from a volumetric map, our method guides the MAV away from both collision points and observed obstacles, enhancing its ability to avoid further collisions and maintain safe navigation. It is important to note that the volumetric map used for collision recovery control does not consider the collision point cloud, which ensures a timely reaction to move the MAV to a safe position, as the process of cloud point cloud generation and mapping may introduce a slight delay.

Firstly, we need to get the collision recovery position $\bm{p}_r$ in the body frame as follows:
\begin{equation}
    {\bm{p}}_{r} = (w\frac{\bm{G}} {||\bm{G}||} - (1-w)\hat{\bm{C}}) R_d
\end{equation}
In this equation, the weight $w$ is calculated by:
\begin{equation}
    w = \left\{
    \begin{array}{ll}
    \frac   {l^3(D - D^*)}
            {D^3(l - D^*)}, 
            & \text{if } l < D \le D^* \\
    0, & \text{otherwise}
    \end{array}
    \right.
\end{equation}
where $\bm{G} = \nabla{D(\bm p_n)}$ represents the gradient vector of the Euclidean Distance Transform (EDT) value at the drone's current position $\bm p_n$ within a local volumetric map. This gradient vector points the direction of the fastest increase in distance to the nearest obstacle. Meanwhile, $D = D(\bm p_n)$ is the EDT value at the current position of the MAV, indicating the shortest distance to the nearest obstacle. The details of the construction of a volumetric map will be discussed in Section \ref{Collision-aware Volumetric Mapping}. When calculating the weight, the lower bound for $D$ is set as the radius $l$ of the drone cage because the closest obstacle should be outside the cage. $D^*$ is an empirically determined upper bound according to the target scenario. Additionally, the reaction distance $R_d = 2l$ constrains the reaction position to a circle with a diameter equal to that of the MAV, which ensures sufficient safe space and enables effective operation in confined environments.

Then, the collision recovery position in the world frame is denoted by ${^{w}{\bm{p}}_r} = {^{w}\bm{T}_{b}} \cdot {{\bm{p}}_r}$, which utilizes a transform matrix ${^{w}\bm{T}_{b}}$ to achieve the transformation from the body frame to world frame. Instead of relying on motion planning, which typically requires re-planning time, the position command is sent directly to the low-level controller with PX4 firmware \cite{meier2015px4}. In the autopilot, a cascaded proportional–integral–derivative (PID) controller is used to generate thrust force commands from the desired reaction positions.



\subsection{Collision Point Cloud Generation}

The collision point cloud generation module is designed to record the positions of unobserved obstacles and avoid a secondary collision. This module constructs a set of points to fit the collision plane where the drone detects the collision. The collision point cloud is then registered in the global volumetric map for the motion planning algorithm to avoid invisible obstacles when re-planning the feasible path.

Firstly, we posit that the object colliding with the drone's surface forms a circular plane with a radius $r_c$ and center point ${\bm{p}_0}$ under the body frame. Here, $r_c$ is defined as the minimum enclosing circle of the MAV. The center point $\bm{p}_0 = [p_{x,0}, p_{y,0},p_{z,0}]^T$ corresponds to the collision point ${\bm{c}_{c}}$, as determined by the collision estimation module. The 3D collision circular plane can be constructed as an intersection of a sphere and a plane, as described by Equation \eqref{eq:3D circle}. Notably, the plane Equation \eqref{eq:3D circle} is defined by the point $\bm{p}_0$ and a normal vector $ \bm{n} = \bm{p}_0 - \bm{c}_0$, where $\bm{c}_0 = [0,0,0]^{\top}$ represents the center point of the MAV in the body frame.
\begin{gather}
\left\{
\begin{aligned}
       \left(p_x-p_{x,0}\right)^2 &+\left(p_y-p_{y,0}\right)^2+\left(p_z-p_{z,0}\right)^2 \leq r_c^2 \\ 
        &\bm{n}^{\top} \begin{bmatrix} p_x-p_{x,0}\\ p_y-p_{y,0}\\ p_z-p_{z,0} \end{bmatrix}  = 0  
\end{aligned}
\right.
\label{eq:3D circle}
\end{gather}

%
Guided by Equation \eqref{eq:3D circle}, a sphere point cloud, denoted as $P_{sph}$, can be generated using point cloud library (PCL). Each point in $P_{sph}$ is then evaluated to determine if it satisfies the plane fitting condition. The points meeting this condition are selected to construct the 3D collision circular plane point cloud, $P_{cir}$, in the body frame. This point cloud is subsequently transformed to the world frame using $^{w}P_{cir} = {^{w}\bm{T}_{b}} \cdot {P_{cir}}$ for building a collision-aware map.

\subsection{Collision-aware Volumetric Mapping}
\label{Collision-aware Volumetric Mapping}

For autonomous navigation purposes, we represent the environment with the help of a volumetric mapper \cite{chen2022gpu}. The mapping system constructs Occupancy Grid Maps (OGMs) and Euclidean Distance Transforms (EDTs) by parallel computing in GPU. An OGM contains the probability of a voxel (an element of the 3D grid) being occupied by obstacles, while an EDT consists of structural voxel grids where every voxel contains the distance information to its closest obstacle. 

The mapper reads the input data of depth and poses from onboard sensors and constructs OGM incrementally. Within the local range, a parallel EDT algorithm converts a batch of OGM in the local volume to EDT. In detail, given a 3D voxel $v$, the distance value $E$ is computed in the way 
\begin{gather}
    E(v) = {\min_{u\in O}||u - v||}
\end{gather}
where $O$ denotes the set of voxels that are occupied.
Finally, the new observation in the local range is integrated into the global map. The actual distance value is propagated outside the local range by parallel wavefront algorithms \cite{chen2022gpu}, and
the global EDT can be obtained. After the construction of OGM and EDT, voxels in the map are labeled in three states, \textit{occupied}, \textit{free}, and \textit{unknown}. Besides, each observed voxel records its distance from the closest obstacle.  Hence, the motion planner will drive the vehicle towards the goal through the observed region while avoiding occupied grids.

We specially tailor the volumetric mapper for Air Bumper. The collision detection mechanism is modeled as a sensor that generates observations of an obstacle, which we refer to as a \textit{collision sensor} in below.
Upon receiving the point cloud from a collision sensor, the mapper uses a feature extractor from PCL to encapsulate all points to an OBB (oriented bounding box). The bounding vertices and corresponding transformation matrix associated with each collision-induced OBB are stored in the mapper and further streamed to GPU in  OGM updating stage. After the local OGM is constructed with onboard sensor observation, the mapper inspects each voxel in parallel to check if the corresponding voxel should be set as occupied in the global OGM. In a thread dealing with the voxel $v$,  all OBBs are iterated, and $v$ is transformed into each OBB coordinate.  If $v$ is inside one of the OBBs marked by the collision sensor, or it is  \textit{occupied}  in the local OGM, then the global OGM increases the occupancy probability of $v$. This indicates the collision sensor has a higher priority than onboard sensors, in that the obstacle registered by the collision sensor will not be cleared by onboard sensors. Local OGM is updated accordingly, and EDT takes the observation of the collision sensor as well. In consequence, the vehicle remembers all obstacles it ever collides with and will avoid them in future navigation.

\subsection{Collision Reaction Motion Planning}
Once the MAV detects the collision, the motion planning model will re-plan the trajectory based on the updated collision-aware OGM and EDT. Here, we employ our previously developed GTO-MPC algorithm \cite{xi2021gto} to plan a smooth trajectory that simultaneously avoids obstacles and achieves the pre-collision goal state $\mathbf{x}_g$. Here, $\mathbf{x}$ represents a state vector, which contains position, velocity, and acceleration in the world frame, following the notation in \cite{xi2021gto}.
This algorithm is divided into two steps. 
Firstly, a jerk-limited trajectory discrete by a series state $\mathbf{x}_j$ is generated using the given goal state $\mathbf{x}_g$ and current state $\mathbf{x}$ to supply the guiding time-optimal (GTO) initial solution. Subsequently, an MPC-based method is employed to follow this trajectory, taking into account both obstacle avoidance and dynamic constraints.
Therefore, for each replanning horizon $t \in [t_0,t_0+T]$, the problem can be formulated as:
\begin{equation}
\begin{array}{ll}
\min \quad J & =\int_{t_0}^{t_0+T} \mathbf{u}^2(t) d t+w_1 \int_{t_0}^{t_0+T}\left\|\mathbf{x}(t)-\mathbf{x}_j(t)\right\|^2 d t \\
& +w_2 \int_{t_0}^{t_0+T} e^{-\left\|d(t)\right\|} d t \\
\end{array}
\label{eq: motion planning}
\end{equation}
where the first term of $J$ minimizes the control input $\mathbf{u}(t)$, which corresponds to the jerk (the derivative of acceleration). This term encourages the smoothness of the trajectory. The second term is to minimize the errors between the state trajectory $\mathbf{x}(t)$ and jerk limited trajectory $\mathbf{x}_j(t)$. The third term penalizes the closest distance, denoted as $d(t)$, from the drone to the nearest obstacles. The distance information is obtained from the collision-aware EDT map. 



\section{Experiments And Results}
\label{Experimental Results}

\subsection{Simulation in an Unknown Environment}

\begin{figure}[h]
    \centering
    \includegraphics[width=1.0 \linewidth]{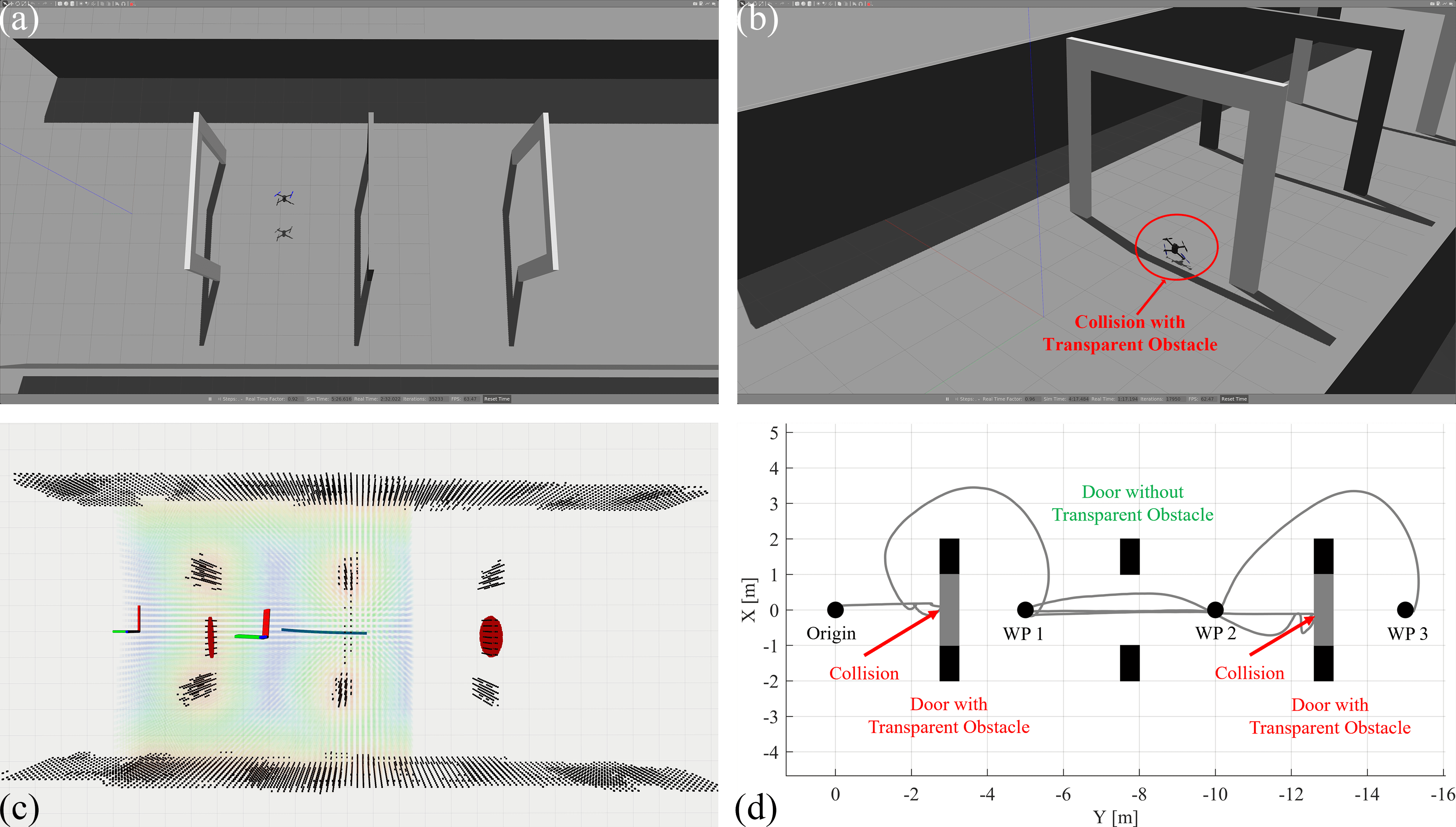}
    \vspace{-18pt}
    \caption{MAV with Air Bumper successfully detects and recovers from collisions with two transparent obstacles in a simulation environment. (a) Overview of the customized environment. (b) MAV, without our framework, crashed due to the collision with a transparent obstacle. (c) Illustration of the collision-aware map with collision point cloud for the simulation. (d) The trajectory of the simulated MAV with Air Bumper framework.}
    \label{fig: simulation}
    \vspace{-6pt}
\end{figure}

We use a customized environment to evaluate the Air Bumper framework in the Gazebo \cite{koenig2004design} simulator, as shown in Fig. \ref{fig: simulation}(a). In the customized environment, we use a simulated MAV with a Velodyne VLP-16 LiDAR sensor, which has $360\degree$ horizontal FOV, $30\degree$ vertical FOV, and the maximum sensing distance is 100m. Two kinds of doors are designed to validate the framework. One is a black door frame without any obstacles. The other one is a white door frame and transparent material, like glass, within the frame, and it is used to simulate a scenario with the aforementioned transparent obstacles. LiDAR is unable to detect transparent obstacles during flight. As a result, the motion planning module may generate a path from the current position to the next goal that passes through the white glass door. This could cause the MAV, without our framework, to become stuck or crash.

In the simulation test, we set three doors: two white doors with transparent obstacles located at $[0, -3, 1]^\top$ m and $[0, -8, 1]^\top$ m, and one black normal door at $[0, -13, 1]^\top$ m. Once the start command is received, the drone takes off and flies autonomously through waypoints (WPs). It follows a path from the origin point $[0, 0, 1]^\top$ m to the first waypoint (WP1) $[0, -5, 1]^\top$ m, then to the second waypoint (WP2) $[0, -10, 1]^\top$ m, and finally to the third waypoint (WP3) $[0, -15, 1]^\top$ m. Without our collision detection and reaction framework, the MAV collides with the transparent obstacles and crashes when passing through the white glass doors (Fig. \ref{fig: simulation}(b)). In contrast, our Air Bumper framework enables the MAV rapidly recover from the collision upon detecting the abnormal acceleration data in the $y$ direction. The collision-aware mapping module consequently updates the collision-aware map, where estimated obstacles are marked in red in Fig. \ref{fig: simulation}(c). The collision-aware map assists the motion planning module in re-planning a smooth trajectory to the goal without colliding with the same obstacles (Fig. \ref{fig: simulation}(d)). Results demonstrate that our framework is able to handle several collisions with unobserved obstacles during autonomous flight and record the collision for further safe navigation.

\subsection{Experiments in Real World}

\begin{figure}[h]
    \vspace{-6pt}
    \includegraphics[width=1.0 \linewidth]{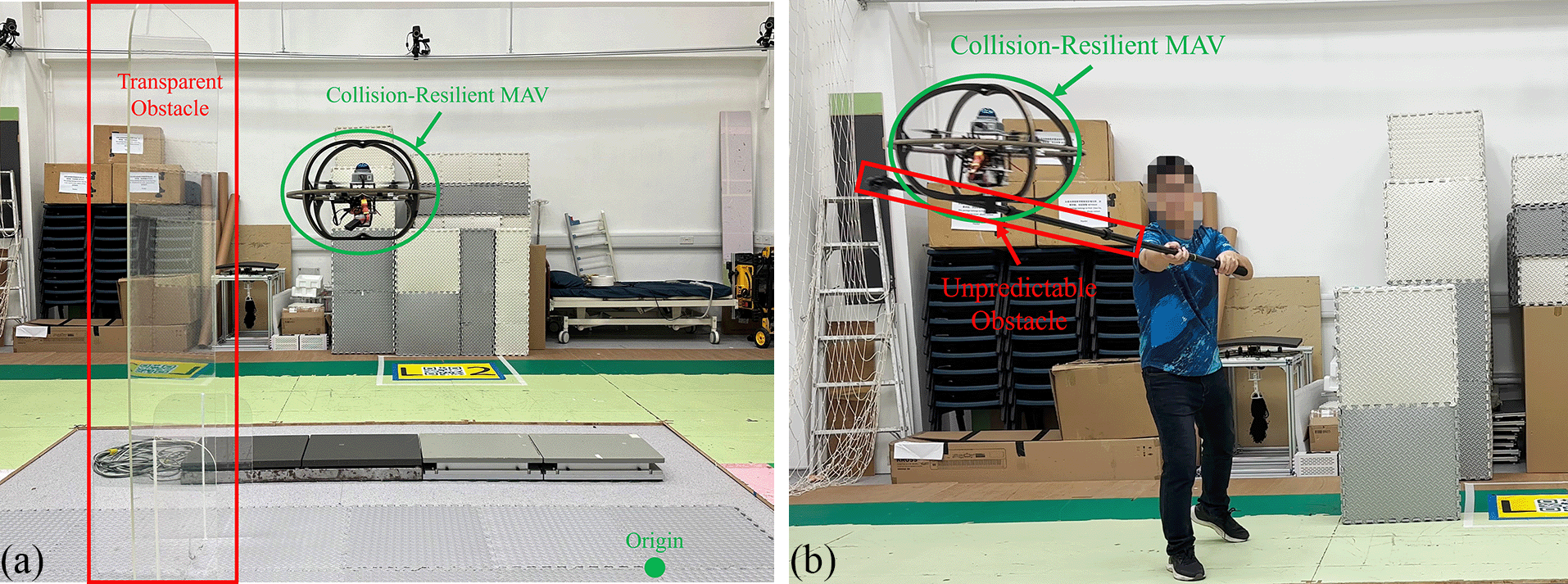}
    \vspace{-18pt}
    \caption{The snapshots of testing Air Bumper in the unknown environment with (a) transparent and (b) unpredictable obstacles.}
    \label{fig:real_scenario}
    \vspace{-6pt}
\end{figure}

The Air Bumper framework's performance is demonstrated using the collision-resilient MAV designed in Section \ref{Design of Collision-Resilient MAV} in an unknown indoor environment with a transparent obstacle (Fig. \ref{fig:real_scenario}(a)) and an unpredictable obstacle (Fig. \ref{fig:real_scenario}(b)). The MAV is programmed to autonomously take off from the origin to the first waypoint (WP1) $[0.0, 0.0, 1.5]^\top$ m, then fly towards the second waypoint (WP2) $[0.0, -3.5, 1.5]^\top$ m, and then perform back-and-forth flights between the two waypoints.

\begin{figure}[htbp]
    \centering
    \vspace{-6pt}
    \includegraphics[width=1.0 \linewidth]{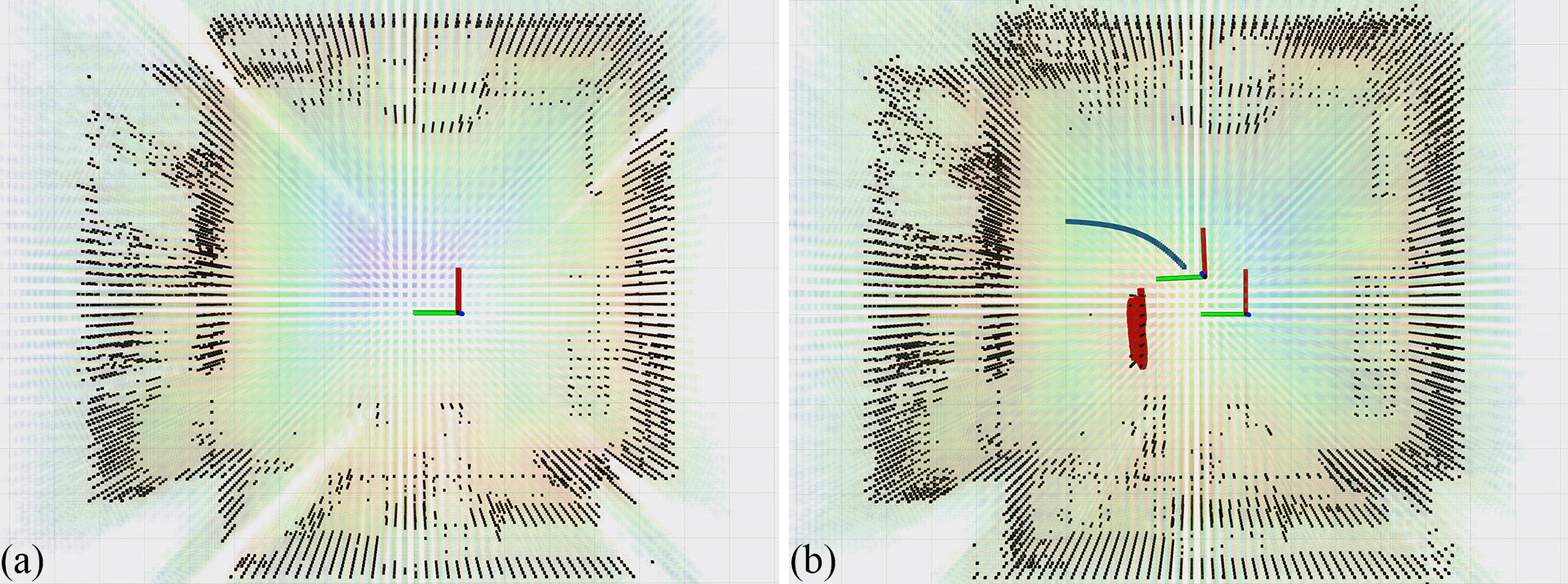}
    \vspace{-18pt}
    \caption{The collision-aware map (a) before and (b) after a collision.}
    \label{fig:real_map}
    \vspace{-6pt}
\end{figure}

For the scenario with a transparent obstacle, a customized transparent object with a size of $2 \times 1$ m and a thickness of 8 mm is considered an obstacle. The bottom center of the obstacle is located at $[0.0, 1.7, 0.0]^\top$ m. The OGM in Fig. \ref{fig:real_map}(a), represented by the black point cloud, demonstrates that the laser beams are able to penetrate the transparent object. Therefore, there are no occupied voxels in the proximity of the obstacle's location, and the motion planning algorithm plans a path through the obstacle, which leads the MAV to collide with the transparent obstacle. 

\begin{figure}[htbp]
    \vspace{-6pt}
    \centering
    \includegraphics[width=1.0 \linewidth]{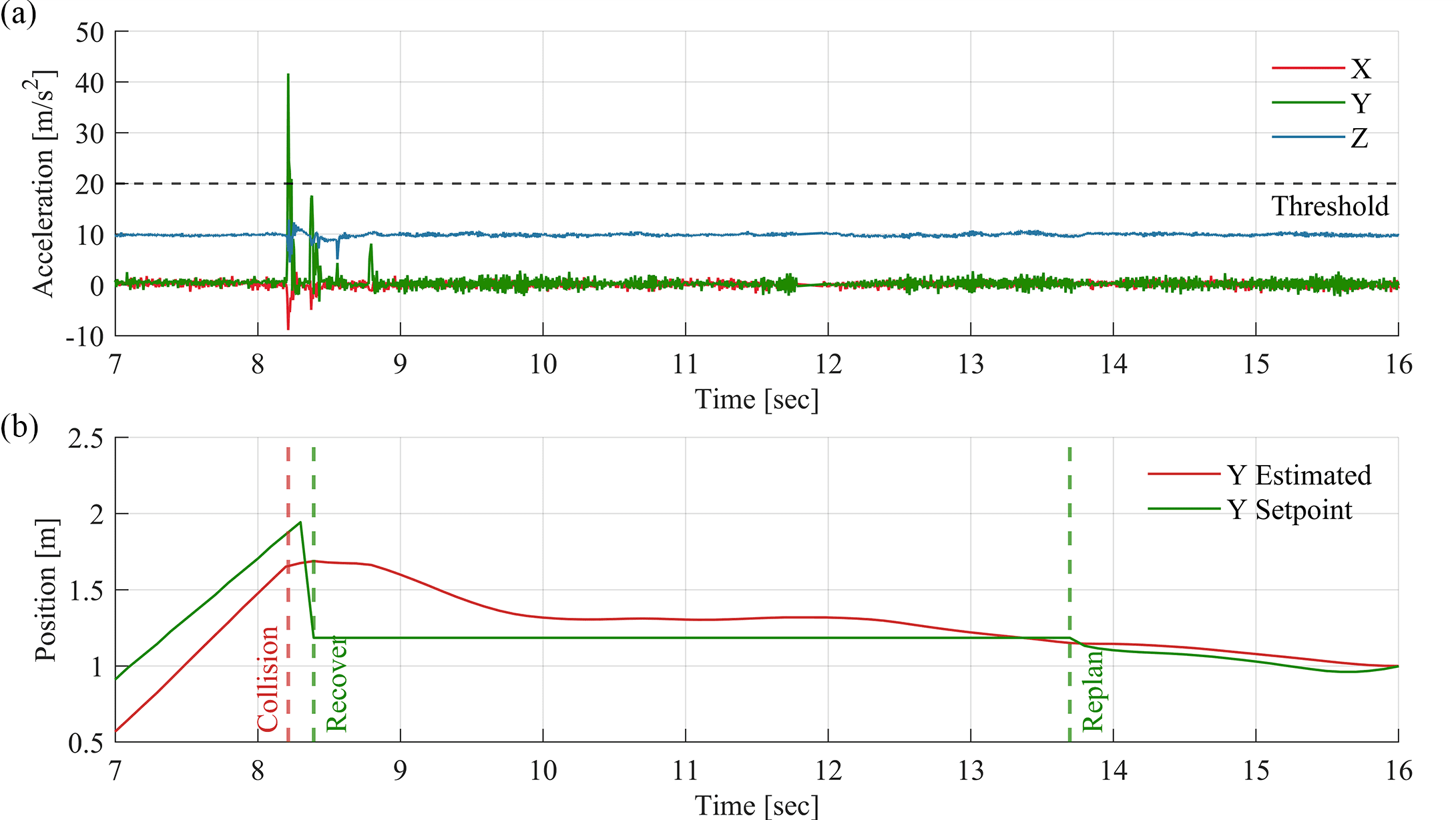}
    \vspace{-18pt}
    \caption{States of the MAV before and after a collision with a transparent obstacle. (a) During the collision, acceleration on $y$-direction exceeds the detection threshold. (b) Position estimation and setpoints on $y$-direction.}
    \label{fig:real_y}
    \vspace{-6pt}
\end{figure}

In one of the flight tests, the collision generates an abnormal acceleration on the $y$-axis, exceeding the threshold, which occurred at approximately 8.21 seconds (Fig. \ref{fig:real_y}(a)). The collision is detected and estimated as $\bm{C}$ with $\varphi = 102.1\degree$ and $\theta = 94.2\degree$. Then, the collision recovery control module calculates and generates a recovery position setpoint in the negative $y$-direction to move the drone away from the obstacle at around 8.39 seconds (Fig. \ref{fig:real_y}(b)). The recovery position ensures the drone is at a safe distance of approximately 0.46 meters from the obstacle. Meanwhile, the collision-aware map is updated after receiving the collision point cloud, marked red in Fig. \ref{fig:real_map}(b). With the help of the collision-aware map, GTO-MPC re-plans a feasible trajectory to the second waypoint, which is shown as a blue line in Fig. \ref{fig:real_map}(b), and the low-level controller executes the re-planned setpoint at around 13.69 seconds (Fig. \ref{fig:real_y}(b)). The framework is designed to allow for a 5-second window after a collision has occurred for the motion planning algorithm to re-plan a feasible path. Nevertheless, the actual time it takes for the drone to recover and stabilize after the collision is less than 1 second.

\begin{figure}[h]
    \centering
    \vspace{-6pt}
    \includegraphics[width=0.75 \linewidth]{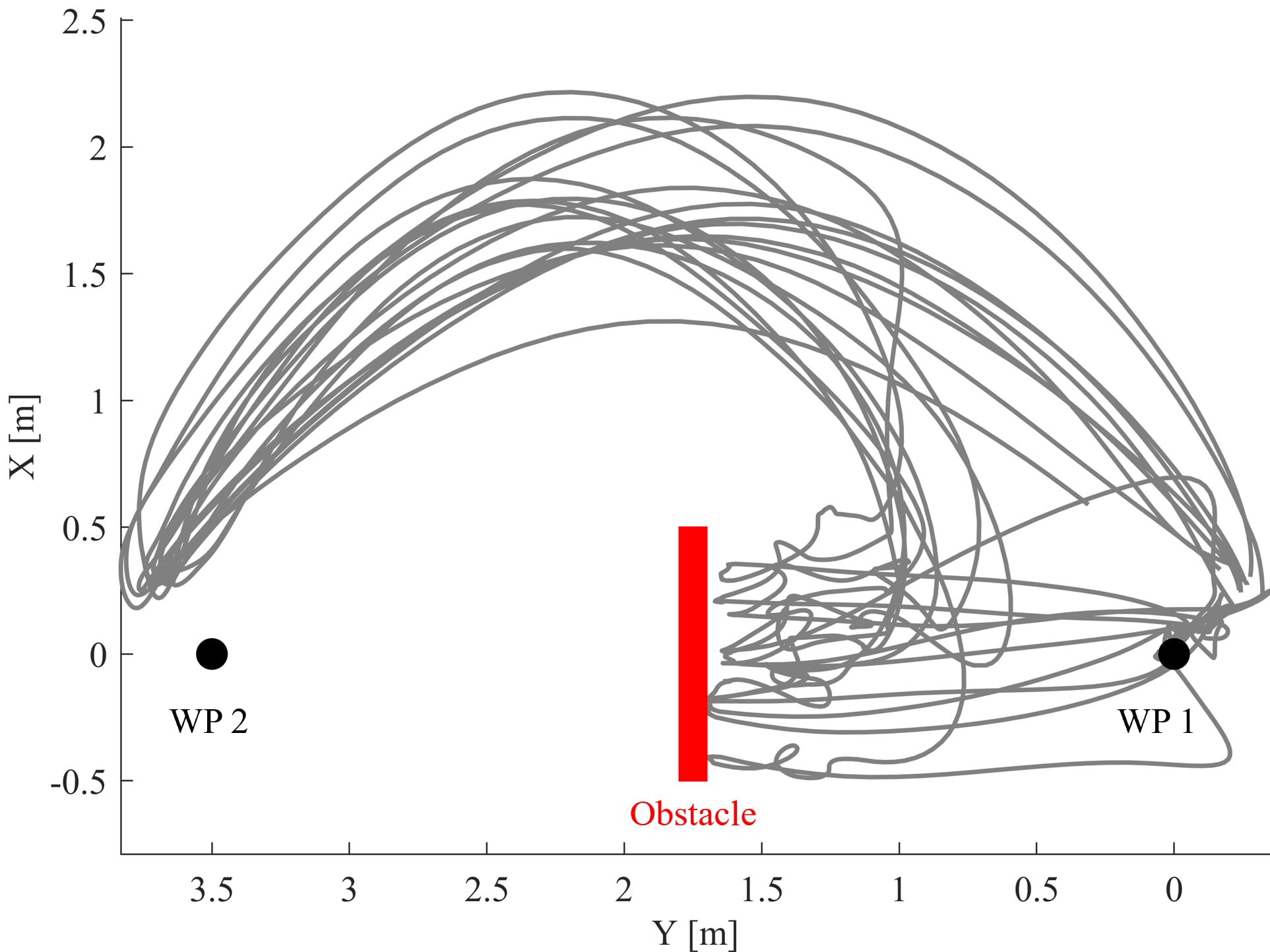}
    \vspace{-6pt}    
    \caption{Real experimental trajectories (ten trails) of the collision-resilient MAV with collision detection and reaction.}
    \label{fig:10times}
    \vspace{-6pt}
\end{figure}

We then conduct ten trials to demonstrate the robustness of our framework for this case. All the experimental trajectories in the testing scenario are shown in Fig. \ref{fig:10times}. In all the trials, the collision-resilient drone collides with the transparent obstacle, and our framework successfully detects and reacts to the collision. Although some of the trajectories do not intersect with the obstacles, the maximum distance between the surface of the obstacle and the collision position is only 0.1 m, which still falls within the drone's radius.

\begin{figure}[!htbp]
    \vspace{-6pt}
    \centering
    \includegraphics[width=1.0 \linewidth]{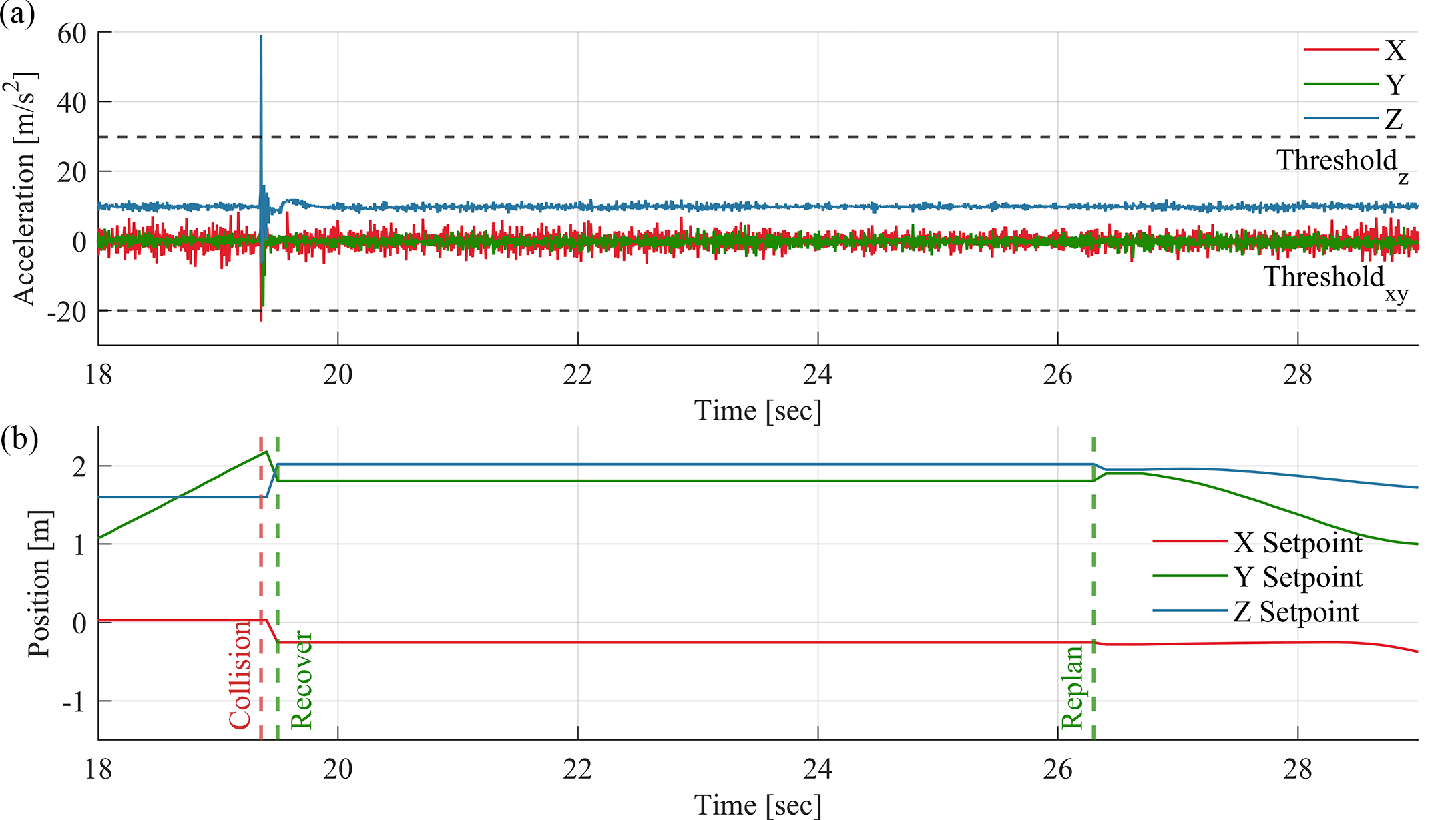}
    \vspace{-18pt}
    \caption{States of the MAV before and after a collision with an unpredictable obstacle. (a) During the collision, accelerations exceed the collision detection threshold. (b) Position setpoints on three axes.}
    \label{fig:real_z}
    \vspace{-6pt}
\end{figure}

For the scenario with an unpredictable obstacle, we use a stick to randomly hit the MAV outside the FOV of the LiDAR to demonstrate the ability of the Air Bumper framework to detect the collision with unpredictable obstacles and perform reactions in 3D space. When the stick hits the MAV from the lower left side of the cage, there are abnormal acceleration data on all three axes (Fig. \ref{fig:real_z}(a)), and the collision is detected at 19.36 s. Then a 3D recovery control is performed with setpoints on three axes at 19.49 s (Fig. \ref{fig:real_z}(b)), the recovery distance $R_d$ ensures the drone is at a safe distance of approximately 0.46 m from the obstacle in the $xy$-plane and makes the drone ascend from about 1.5 m to about 2 m along $z$-axis. Results demonstrate that our framework enables the MAV to maintain a safe distance from the obstacle in 3D space rather than just in a certain plane. Then the collision-aware mapping module generates a collision point cloud at the collision point, which helps the motion planning module generate a smooth feasible trajectory to the pre-collision goal state successfully.


\section{Conclusion}
\label{Conclusion}

In this work, we introduced a collision detection and reaction framework to help MAVs recover from collisions during autonomous flights in an unknown environment with unobserved obstacles. To do so, we designed an IMU-based collision detection and estimation module to estimate the collision intensity, direction, and position. Collision reaction modules are developed to assist the drone quickly away from the obstacle and update the collision-aware map to generate a smooth post-collision trajectory. In addition to the software, a caged collision-resilient MAV is also designed and crafted, which fully demonstrates the ability of our framework in the real world. The motion planning algorithm in the current framework still needs a certain time to re-plan after collisions. In the future, we aim to introduce collision-inclusive motion planning, which can better utilize collisions in autonomous navigation in complex environments. The framework can be further extended to assist multi-robot navigation in hazardous environments.






\bibliographystyle{conf/IEEEtran}
\balance
\bibliography{reference}


\end{document}